\documentclass[11pt,a4paper]{article}
\usepackage[hyperref]{eacl2021}
\usepackage{times}
\usepackage{latexsym}

\usepackage{url}
\usepackage{latexsym}
\usepackage{graphicx}
\usepackage{amssymb, amsfonts, amsmath}
\usepackage{multicol}
\usepackage[normalem]{ulem}
\usepackage{multirow}
\usepackage{graphicx}
\usepackage{bbm}
\usepackage{colortbl}
\usepackage{wrapfig}
\usepackage{bm}
\usepackage[russian,english]{babel}[T2A]
\usepackage{caption}
\usepackage{subcaption}
\usepackage{xcolor}

\usepackage{microtype}

\aclfinalcopy % Uncomment this line for the final submission
\def\aclpaperid{864} %  Enter the acl Paper ID here

\newcommand{\R}{\mathbb{R}}

\title{A Unified Feature Representation for Lexical Connotations}

\author{Emily Allaway\\
  Columbia University  \\
    New York, USA \\
  \texttt{eallaway@cs.columbia.edu} \\
  \And
  Kathleen McKeown \\
Columbia University  \\
    New York, USA \\
  \texttt{kathy@cs.columbia.edu} \\}

\date{}

\begin{document}
\maketitle

\begin{abstract}
Ideological attitudes and stance are often expressed through subtle meanings of words and phrases.
Understanding these {\em connotations} is critical to recognizing 
the cultural and emotional perspectives
of the speaker. In this paper, we use distant labeling to create a new lexical resource representing  connotation aspects for nouns and adjectives. Our analysis shows that it aligns well with human judgments. Additionally, we present a method for creating lexical representations that capture connotations within the embedding space and show that using the embeddings provides a statistically significant improvement on the task of stance detection when data is limited.
\end{abstract}

\section{Introduction}
Expressions of ideological attitudes are widespread in today's online world, influencing how we perceive and react to events and people on a daily basis.
These attitudes are often expressed through subtle expressions or associations~\citep{Somasundaran2010RecognizingSI,Murakami2010SupportOO}. For example, the sentence ``the people opposed gun control'' conveys no information about the author's opinion. However, by adding just one word, ``the \textit{selfish} people opposed gun control'', the author can convey 
their stance
on both gun control (against) and the people who support it (not valuable and disliked). Discerning such subtle meaning is crucial for fully understanding and recognizing the hidden influences behind 
everyday content.
% the barrage of content we encounter every day.  

Recent studies in NLP have begun to examine these hidden influences through framing in social media and news~\citep{Asur2010PredictingTF,Hartmann2019IssueFI,Klenner2017AnOM} and style detection in hyperpartisan news~\citep{Potthast2017ASI}. 
Lexical
connotations provide a method to study these influences, including stance, in more detail.

\begin{figure}[t]
    \centering
    \includegraphics[width=.42\textwidth]{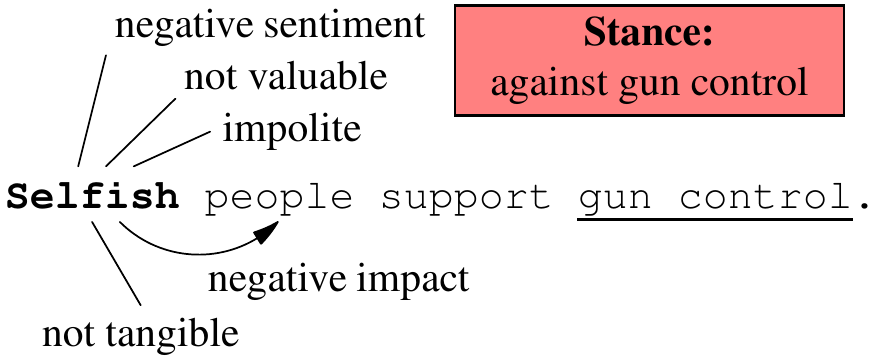}
    \caption{Connotations of the word ``selfish'' and the resulting implied stance 
    % (opinion) 
    on the topic ``gun control''.}
    \label{fig:intro}
    \vspace{-10pt}
\end{figure}
Connotations are implied cultural and emotional associations for words that augment their literal meanings
~\citep{Carpuat2015ConnotationIT,Feng2011LearningGC}.
Connotation values are \textit{associated} with a phrase (e.g., fear is associated with ``cancer'')~\citep{Feng2011LearningGC}
and capture a range of nuances, such as whether a phrase is an insult or 
% whether it 
implies value (see Figure~\ref{fig:intro}).

In this paper, we define six new fine-grained connotation aspects for nouns and adjectives, filling a gap in the
%KM adding commas around which
literature on connotation lexica, which has focused on verbs ~\citep{Sap2017ConnotationFO,rashkin-etal-2016-connotation,rashkin-etal-2017-multilingual}, and coarse-grained polarity ~\citep{Feng2011LearningGC,kang-etal-2014-connotationwordnet}. 
We create a new distantly labeled English lexicon that maps nouns and adjectives to our six aspects
% . We then show that 
and show that it
% our lexicon
aligns well with human judgments.
In addition, we show that our lexicon confirms existing hypotheses about subtle semantic differences between synonyms.

We then learn a single connotation embedding space for words from all parts of speech, combining our lexicon with existing verb lexica and contributing to the literature on unifying lexica~\citep{Hoyle2019CombiningSL}.
Intrinsic evaluation shows that our embedding space captures clusters of connotatively-similar words. In addition, our embedding model can generate representations for new words without the numerous training examples required by standard word-embedding methods.
Finally, we show that our connotation embeddings improve performance on stance detection, particularly in a low-resource setting. 

Our contributions are as follows: (1) we create a new connotation lexicon and show that it aligns well with human judgments, (2) we train a connotation feature embedding for all parts of speech and  show that it captures connotations within the embedding space, and (3) we show the connotation embeddings improve stance detection when data is limited. Our resources are available: \url{https://github.com/emilyallaway/connotation-embedding}.
% new resources will be made available.

\section{Related Work}
Studies of connotation build upon the literature examining subtle language nuances, including good and bad effects of verbs~\citep{Choi2014EffectWordNetSL}, evoked sentiments and emotions~\citep{mohammad-etal-2013-nrc,Mohammad2010EmotionsEB,Mohammad2018WAI}, multi-dimensional sentiment~\citep{Whissell2009UsingTR,Mohammad2018VADlex,Whissell1989THEDO}, offensiveness~\citep{Klenner2018OffensiveLW}, 
and 
psycho-sociological properties of words~\citep{Stone1963ACA,Pennebaker2001LIWCLI}. Work explicitly on connotations has focused primarily on detailed aspects for verbs~\citep{rashkin-etal-2016-connotation,rashkin-etal-2017-multilingual,Sap2017ConnotationFO,Klenner2017AnOM} or single polarities for many parts of speech~\cite{Feng2011LearningGC,feng-etal-2013-connotation,kang-etal-2014-connotationwordnet}. One exception is the work of ~\citet{Field2019ContextualAA}, which extends limited detailed connotation dimensions from verbs to nouns within the context of certain verbs. 
Our work is unique in 
directly defining detailed aspects for nouns and adjectives. 

Early work on stance detection applied topic-specific models to various genres, including online debate forums~\citep{sridhar-etal-2015-joint,Somasundaran2010RecognizingSI,Murakami2010SupportOO,hasan-ng-2013-stance,Hasan2014WhyAY} and student essays~\citep{Faulkner2014AutomatedCO}. More recent studies have used a single model for many topics to predict stance in Tweets~\citep{Mohammad2016SemEval2016T6,Augenstein2016StanceDW,Xu2018CrossTargetSC}
and as part of the fact extraction and verification pipeline~\citep{Conforti2018TowardsAF,Ghanem2018StanceDI,Riedel2017ASB,hanselowski-etal-2018-retrospective}.
\citet{klenner-etal-2017-stance} explore the relationship between connotations and stance through verb frames. In contrast, our work studies stance 
using connotation representations from a learned joint embedding space for words from all parts of speech. Recently,~\citet{Webson2020Conn} examine representations of political ideology as connotations and its use in information retrieval.
Representation learning has been used for stance detection of online
debates by~\citet{li-etal-2018-structured}, who develop a joint representation of the text and the authors. Our work, however, uses a representation of word connotations and does not use any author information (a strong feature in fully-supervised datasets but which may not be available in real-world settings).

\section{Connotation Lexicon}
\label{sec:lex}
We build a connotation lexicon for nouns and adjectives by defining six new aspects of connotation.
We take inspiration from verb connotation frames and their extensions~\citep{rashkin-etal-2016-connotation,Sap2017ConnotationFO}, which define aspects of connotation in terms of the agent and theme of transitive verbs. \citet{rashkin-etal-2016-connotation} define six aspects of connotation for verbs (entities', writer's, and reader's perspectives, effect, value, and mental state) in  connotation frames (e.g., ``suffer'' $\leadsto$ negative effect on the agent) and \citet{Sap2017ConnotationFO} extend these aspects to include power and agency. 

We first define the six new aspects of connotation for nouns and adjectives (\S \ref{sec:conndefs}) in our work, then we describe our distant labeling procedure (\S \ref{sec:labels}) and human evaluation of the final lexicon (\S \ref{sec:conneval}).

\subsection{Definitions}
\label{sec:conndefs}
We use $w$ to indicate a word and $w'$ to indicate the person, thing or attribute signified by $w$.

For each $w$, we define (1) \textbf{Social Value}: whether  $w'$ is considered valuable by society, (2) \textbf{Politeness} \textit{(Polite)}: whether $w$ is a socially polite term, (3) \textbf{Impact} whether $w'$ has an impact on society (or the thing modified by $w$ if $w$ is an adjective), (4) \textbf{Factuality} \textit{(Fact)}: whether $w'$ is tangible, (5) \textbf{Sentiment} \textit{(Sent)}: the sentiment polarity of $w$, and (6) \textbf{Emotional association} \textit{(Emo)}: the emotions associated with $w'$. We show examples in Table~\ref{tab:rules}.

\begin{table*}[t]
\centering
\begin{tabular}{llll}
\hline
\textbf{Aspect}& \textbf{Lexicon} & \textbf{Example Rules} & \textbf{Examples} \\ \hline
\begin{tabular}[c]{@{}l@{}} \textit{Social}\\ \textit{Value} \end{tabular}& GI &\begin{tabular}[c]{@{}l@{}} Authoritative power $\rightarrow$ valuable \\ Related to failure $\rightarrow$ not valuable\end{tabular} & \begin{tabular}[c]{@{}l@{}} \texttt{attorney} $\leadsto$ valuable (+) \\ \texttt{aimless} $\leadsto$ not valuable (-) \end{tabular} \\ \hline
\textit{Politeness} 
% \textit{Polite}
& GI &\begin{tabular}[c]{@{}l@{}} Gain of respect $\rightarrow$ polite\\ Loss of affection $\rightarrow$ impolite\end{tabular}  & \begin{tabular}[c]{@{}l@{}} \texttt{commendable} $\leadsto$ polite (+) \\ \texttt{alienation} $\leadsto$ impolite (-) \end{tabular} \\ \hline
\textit{Impact}& GI &\begin{tabular}[c]{@{}l@{}} Virtue $\rightarrow$ positive\\ Loss of well-being $\rightarrow$ negative \end{tabular} & \begin{tabular}[c]{@{}l@{}}\texttt{adept} $\leadsto$ positive impact (+)\\ \texttt{shock} $\leadsto$ negative impact (-) \end{tabular} \\\hline
\textit{Factuality}
% \textit{Fact}
& DAL &\begin{tabular}[c]{@{}l@{}} Imagery($w$) $> \theta_F$ $\rightarrow$ factual\\ Imagery($w$) $< -\theta_F$ $\rightarrow$ not factual \end{tabular} & \begin{tabular}[c]{@{}l@{}} \texttt{rocky} $\leadsto$ factual (+) \\\texttt{tradition} $\leadsto$ not factual(-)  \end{tabular} \\\hline
\textit{Sentiment}
% \textit{Sent}
& CWN &\begin{tabular}[c]{@{}l@{}}  $v > \theta_S \rightarrow$ positive\\  $v < -\theta_S \rightarrow$ negative \end{tabular} & \begin{tabular}[c]{@{}l@{}} \texttt{song} $\leadsto$ positive (+) \\ \texttt{cancerous} $\leadsto$ negative (-) \end{tabular} \\\hline
\begin{tabular}[c]{@{}l@{}} \textit{Emotional}\\ \textit{Association} \end{tabular} 
% \textit{Emo}
& NRC &\begin{tabular}[c]{@{}l@{}}  emotions $E \subseteq \{$anger, joy, fear, trust,\\  anticipation, sadness, disgust, surprise$\}$ \end{tabular} & \begin{tabular}[c]{@{}l@{}} \texttt{snake} $\leadsto$ $\{$disgust, fear$\}$ \\ \texttt{effective} $\leadsto$ $\{$trust$\}$ \end{tabular} \\\hline
\end{tabular}
\caption{Example mappings from existing lexica to our connotation aspects. GI: Harvard General Inquirer, DAL: Dictionary of Affect in Language, CWN: Connotation WordNet, and NRC: NRC Emotion Lexicon. Scores for imagery, Imagery($w$), and sentiment, $v$, are real-valued.}
\label{tab:rules}
% \vspace{-10pt}
\end{table*}
\textbf{(1) Social Value} includes both the value of objects or concepts and the social status and power of people or people-referring nouns (e.g., occupations).
``Sociocultural pragmatic reasoning''~\citep{ColstonKatz} about such factors
is crucial for understanding 
% non-literal 
language such as connotations.

Initial work on connotation polarity lexica recognized the important role of \textit{Social Value} in overall connotation by defining a `positive' connotation for objects and concepts that people value~\citep{Feng2011LearningGC}. Later work made this idea more explicit by defining `Value' for transitive verb arguments in connotation frames. More recently `power' and `agency', components of \textit{Social Value}, have been defined for verbs in connotation frames and for nouns in context~\citep{Field2019ContextualAA} and have been used to analyze bias and framing in a variety of texts, illustrating the applications and importance of \textit{Social Value} in connotations.

\textbf{(2) Politeness} follows the definition of Lakoff~\shortcite{LakoffPolite} in noting words that make the addressee feel good but also includes
notions of formality. These notions have been previously studied within the context of politeness as a set of behaviors and linguistic cues~\citep{BrownLevinson,danescu-niculescu-mizil-etal-2013-computational,Aubakirova2016InterpretingNN}. 
We focus on purely lexical distinctions because
%KMMay31 - is "comprehending" the right word here? Or is it grammatically correct? There is no other subject so the implication is "we comprehend". Not sure that's what  you meant. Maybe you mean "because how a reader comprehends these ...."?
% comprehending 
how one comprehends
these distinctions affects one's ``attitude towards the speaker ... or some issue'' as well as whether one feels insulted by the exchange~\citep{ColstonKatz}.
This aspect of perspective is a component of verb connotation frames and we extend it to nouns and adjectives in our lexicon through \textit{Politeness}.  

\textbf{(3) Impact} and effect have been studied in verb connotation frames and other verb lexica~\citep{Choi2014EffectWordNetSL}, capturing notions of implicit benefit or harm on the arguments of the verb. We extend this idea to nouns and adjectives by observing that while they do not directly have arguments, nouns (e.g. ``democracy'') often impact society and adjectives (e.g. ``sick'') impact
the nouns they modify. Thus, we define \textit{Impact} in this way.

\textbf{(4) Factuality} captures whether words correspond to real-world objects or attributes, following the sense of~\citet{Saur2009FactBankAC}. \citet{Klenner2016HowFD} argue that the factuality of events is crucial for understanding sentiment inferences. Building upon this,
\citet{klenner-etal-2017-stance} use factuality as a key component of German verb connotations and of applying those connotations to analyze stance and sides in German Facebook posts. Imagery, as an ``indicator of abstraction''~\citep{Whissell2009UsingTR}, also models a similar attribute to event factuality for all parts of speech. Given its importance, we include a notion of \textit{Factuality} for nouns and adjectives as aspect of connotations. 

\textbf{(5) Sentiment} polarity has been used to convey overall connotations since the early work on connotation lexica~\citep{Feng2011LearningGC,feng-etal-2013-connotation,kang-etal-2014-connotationwordnet}.
As such, we deem it important to include this polarity in our lexicon.

\textbf{(6) Emotional Associations} for words  can be strong, persisting long after they are formed and improving the recall of memories triggered by those words~\citep{Rubin2006TheBM}. Emotions are also impacted when people process non-literal meaning~\citep{ColstonKatz}. To fully understand what a piece of text is trying to convey, it is important to understand what emotional associations exist in the text. 
For example, news headlines often aim to evoke strong emotions in their readers \citep{Mohammad13}. 
To capture this, we include \textit{Emotional Association} as an aspect of connotation. 
\subsection{Labeling Connotations}
\label{sec:labels}
We use distant labeling to build our lexicon, since complete manual annotation of a lexical resource is a lengthy and costly process. Although crowdsourcing can lessen these burdens, the results are often unreliable with low inter-annotator agreement and, for this reason, many lexical resources are automatically created \citep{mohammad2012STARSEM-SEMEVAL,MohammadKZ2013,kang-etal-2014-connotationwordnet}. Following these 
researchers,
we automatically generate our lexicon by combining several existing lexica.

To generate our lexicon, we map dimensions from existing lexica to connotation aspects (see Table~\ref{tab:rules}). We use dimensions from the Harvard General Inquirer~\citep{Stone1963ACA} for \textit{Social Value}, \textit{Politeness}, and \textit{Impact}.
For \textit{Factuality} we map the real-valued `Imagery' dimension, Imagery($w$), from the revised Dictionary of Affect in Language~\citep{Whissell2009UsingTR} into distinct classes. For \textit{Sentiment} we directly use the polarity $v$ from Connotation WordNet~\citep{kang-etal-2014-connotationwordnet} and for \textit{Emotional Association} we use the eight Plutchik emotions \cite{Plutchikemotions} from the NRC Emotion Lexicon~\citep{Mohammad13} (see appendix~\ref{app:connlabel} for full rules).

The labels are word-sense-independent, 
following other automatically generated lexica, such as the Sentiment140 lexicon \citep{MohammadKZ2013}, which do not treat word sense. In addition, sense-level annotations are not available for all lexica in our distant labeling method and therefore sense-level connotations would require both extensive manual annotation and automated word-sense disambiguation, introducing cost and additional noise. As a result, we use sense-level distinctions (e.g., in the Harvard General Inquirer) when available and combine the labels for an aspect across senses to 
% generate 
obtain
the final connotation aspect label. These aggregate aspect labels represent a word's connotative potential, rather than exact value
% for a word.

\begin{table}[t]
\centering
\begin{tabular}{l|lllll}
    \hline
    &\begin{tabular}[c]{@{}l@{}}\textit{Social}\\ \textit{Val}\end{tabular} & \textit{Polite} & \textit{Impact} & \textit{Fact} & \textit{Sent}\\
    \hline
    $\bm{\%+}$ & 32.1 & 10.5 & 14.8 & 19.0 & 56.8\\
    $\bm{\%-}$ & 15.5 & 1.0 & 13.3 & 67.2 & 33.1\\ \hline
    
    %  & \textbf{\% +} & \textbf{\% $-$ }\\ \hline
    % \textit{Social Value} 
    % & 32.1 & 15.5\\ 
    % \textit{Politeness} & 10.5 & 1.0\\ 
    % \textit{Impact} & 14.8 & 13.3 \\
    % \textit{Factuality} & 19.0 & 67.2\\
    % \textit{Sentiment} & 56.8 & 33.1\\ \hline
\end{tabular}
\caption{Class distributions of fully-labeled words in the connotation lexicon.}
\label{tab:dims}
\end{table}
Our resulting lexicon has $7,578$ words fully-labeled for all aspects, with an additional ${\sim}93$k words labeled only for some aspects (e.g., only \textit{Sentiment}), resulting in $100,176$ words total. For each non-emotion aspect, we have a label $l \in \{-1, 0, 1\}$. For \textit{Emotional Association}, each of the eight emotions has label $l \in \{0,1\}$.

We find that many aspects exhibit uneven class distributions (e.g., $10.5\%$ of words are polite and only $1\%$ are impolite) (see Table~\ref{tab:dims}). For emotions, we calculate the class distribution using the number of fully-labeled words with at least one associated emotion ($1,373$ words or $18\%$). For these $1,373$ words, the average number of associated emotions is ${\sim}2$. Our distributions are similar to previous work on verb connotations, 
where distributions range from $1.4\%$ to $20.2\%$ for the smallest class~\citep{rashkin-etal-2016-connotation}.

\subsection{Evaluation of the Lexical Resource}
\label{sec:conneval}
\noindent
\textbf{Human Evaluation}\\
We evaluate the quality of the lexicon by creating a gold-labeled set and 
comparing the  labels created with distant supervision against the human labels.
We ask nine
NLP researchers\footnote{native English speakers at Columbia University}
to annotate 350 words (175 nouns, 175 adjectives) for \textit{Social Value, Politeness, Impact} and \textit{Factuality}. We do not annotate \textit{Sentiment} or \textit{Emotional Association}, since these labels come directly from existing lexica. 

Annotators are given a word $w$, along with its definitions (for all senses) and related words, and 
annotate connotation independent of word sense.
This setup mimics the input to  
the representation learning models 
in \S \ref{sec:embed}. The average Fleiss' $\kappa$ across nouns and adjectives is $0.60$ (see Table~\ref{fig:humaneval}), indicating substantial agreement. We select as the final annotator label, the majority vote.% of three annotators. 
%KM I inserted a comma above after "label". See if you're OK. 

\begin{table}[t]
\centering
\begin{tabular}{lcccc}
\hline
\textbf{Aspect} & \begin{tabular}[c]{@{}l@{}}\textbf{Avg}\\ \textbf{$\kappa$}\end{tabular} & \begin{tabular}[c]{@{}l@{}}\textbf{Avg \%} \\ \textbf{Agree}\end{tabular} & \begin{tabular}[c]{@{}l@{}}\textbf{Lex \%} \\ \textbf{Agree}\end{tabular} & \begin{tabular}[c]{@{}l@{}}\textbf{Lex \%} \\\textbf{NC}\end{tabular}\\ \hline
\multicolumn{1}{l|}{\begin{tabular}[c]{@{}l@{}}\textit{Social} \\ \textit{Value}\end{tabular}} & .699 & 88.9 & 68.6 & 92.6\\ %\hline
\multicolumn{1}{l|}{\textit{Politeness}} & .381 & 56.6 & 59.4 & 95.1\\ %\hline
\multicolumn{1}{l|}{\textit{Impact}} & .630 & 87.6 & 73.7 & 94.6 \\ %\hline
\multicolumn{1}{l|}{\textit{Factuality}} & .675 & 86.3 & 58.0 & 77.7\\ \hline
\multicolumn{1}{l|}{\textbf{Average}} & .596 & 87.9 & 64.2 & 90.0\\ \hline
\end{tabular}
\caption{Lexicon annotation results. Fleiss' $\kappa$ and \% agreement are averaged across nouns and adjectives. Lex \%  is agreement between annotators and the lexicon. NC indicates non-conflicting value agreement. }\label{fig:humaneval}
\end{table}
\begin{table*}[ht]
\vspace{10pt}
\centering
\begin{tabular}{lll}
\hline
\textbf{Aspect} &  \textbf{Same Connotation} & \textbf{Different Connotation} \\ \hline
\begin{tabular}[c]{@{}l@{}}\textit{Social}\\ \textit{Value}\end{tabular} & \begin{tabular}[c]{@{}l@{}} (=) \texttt{hurry} vs. \texttt{rush}
% hurry vs. rush (=)
\\ (+) \texttt{fantastic} vs. \texttt{wonderful}
% fantastic (+) vs. wonderful (+)
\end{tabular} & \begin{tabular}[c]{@{}l@{}} \texttt{sentence} (=) vs. \texttt{condemnation} (-)\\ \texttt{relentless} (-) vs. \texttt{persistent} (+)\end{tabular} \\ \hline
\textit{Politeness} & \begin{tabular}[c]{@{}l@{}} (-) \texttt{disgrace} vs. \texttt{shame} \\ (+) \texttt{humble} vs. \texttt{modest} \end{tabular} & \begin{tabular}[c]{@{}l@{}}\texttt{gentleman} (+) vs. \texttt{man} (=)\\ \texttt{preposterous} (=) vs. \texttt{ridiculous} (-)\end{tabular} \\ \hline
\textit{Impact} & \begin{tabular}[c]{@{}l@{}}(-) \texttt{weariness} vs. \texttt{fatigue}\\ (+) \texttt{rightful} vs. \texttt{lawful}\end{tabular} & \begin{tabular}[c]{@{}l@{}}\texttt{fire} (=) vs. \texttt{burning} (-)\\ \texttt{supporting} (+) vs. \texttt{suffering} (-)\end{tabular} \\ \hline
\textit{Factuality} & \begin{tabular}[c]{@{}l@{}} (+) \texttt{daytime} vs. \texttt{day}\\ (-) \texttt{dire} vs. \texttt{terrible}\end{tabular} & \begin{tabular}[c]{@{}l@{}}\texttt{post} (+) vs. \texttt{position} (=)\\ \texttt{protection} (-) vs. \texttt{security} (=)\end{tabular} \\ \hline
\textit{Sentiment} & \begin{tabular}[c]{@{}l@{}}(+) \texttt{wonderous} vs. \texttt{marvelous}\\(-) \texttt{commotion} vs. \texttt{disturbance} \end{tabular} & \begin{tabular}[c]{@{}l@{}}\texttt{giddy} (=) vs. \texttt{dizzy} (+)\\ \texttt{moving} (=) vs. \texttt{striking} (+)\end{tabular} \\ \hline
\begin{tabular}[c]{@{}l@{}}\textit{Emotional}\\ \textit{Association}\end{tabular} & \begin{tabular}[c]{@{}l@{}} (trust) \texttt{wise} vs. \texttt{smarter}\\ (sadness) \texttt{flaw} vs. \texttt{disturbance}\end{tabular} & \begin{tabular}[c]{@{}l@{}}\texttt{dire} (fear, sadness) vs. \\\texttt{terrible} (fear,sadness,disgust)\end{tabular} \\ \hline
\end{tabular}
\caption{Synonym examples from the lexicon. = indicates neutral or neither.}\label{tab:synex}
\end{table*}
We find that the distantly labeled lexicon agrees with human annotators
the majority of the time (on average $64.2\%$ or Cohen's $\kappa = 0.368$~\citep{cohen-1988-relationship}). 
If we consider \emph{non-conflicting value agreement} (NC), the lexicon agreement with humans rises to $90\%$, where NC agreement is defined as: the pairs (+, neutral) and ($-$, neutral) agree but (+,$-$) does not.
This shows that the lexicon and humans rarely select opposite values and instead disagree on 
% the distinction of 
neutral vs. non-neutral.

Looking closer at disagreements between neutral and non-neutral, we see that most result from human annotators selecting a non-neutral label. That is, the lexicon makes fewer distinctions between neutral and non-neutral than humans; humans select a non-neutral value $68\%$ of the time, compared to $56\%$ in the lexicon. 
Despite this tendency towards neutral, the lexicon aligns with human 
judgments, agreeing the majority of the time and rarely providing a value opposite to humans.\\

\noindent
\textbf{Synonym Analysis}\\
We also evaluate the ability of our lexicon to capture subtle semantic differences between words using lexical paraphrases (synonyms). 
In the paraphrase literature, it has been argued that paraphrases actually differ in many ways, including in connotations~\citep{Bhagat2013WhatIA}. 
In fact, ~\citet{ClarkConvenCon} proposes that absolute synonymy between linguistic forms does not exist.
With this in mind, we hypothesize that our connotation lexicon should differentiate between lexical paraphrases. 

To test this, we select synonym paraphrase pairs from lexical PPDB \citep{pavlick-etal-2015-ppdb} 
where one element in the pair is in the Wordnet synset 
of the other \footnote{\url{https://wordnet.princeton.edu/}}.
We find that out of the $2216$ resulting pairs where both words are in our lexicon, $74.3\%$ have connotations that differ in some aspect. Many words agree on \textit{Sentiment} ($67.5\%$ the same), following the intuition that two synonyms likely have the same sentiment but differ in more fine-grained ways. Other pairs agree on \textit{Politeness} ($76.1\%$ the same), resulting from the extreme class imbalance for this aspect ($88.5\%$ neutral). However, the lexicon does still capture differences along these dimensions, for example in terms of formality (e.g., ``gentleman'' vs. ``man'').

Looking more closely, we find that many times agreements along a particular dimension accurately represent synonyms that differ along other dimensions. For example, ``weariness'' and ``fatigue'' both have a negative \textit{Impact}, but ``weariness'' is associated with \textit{sadness} and ``fatigue'' is not. 

On the other hand, the majority of differences across almost all aspects ($79\%$ on average) are between neutral and non-neutral polarities within a synonym pair, 
for example, between ``position'' (possibly tangible) and ``post'' (tangible), from \textit{Factuality}. This confirms the intuition that synonyms often do not have opposing connotation values, although examples do exist (e.g., the \textit{Social Value} of ``relentless'' vs. ``persistent'') (see Table \ref{tab:synex}). As a whole, our analysis confirms our hypothesis and the claims of \citet{ClarkConvenCon} about synonymy.

%%%%%%%%%%%%%%%%%%%%%%%%%
% connotation embedding %
%%%%%%%%%%%%%%%%%%%%%%%%%
\section{Connotation Embedding}
\label{sec:embed}
\subsection{Methods}
\label{sec:connmethod}
Using our connotation lexicon, we train a dense connotation feature representation for words from all parts of speech. We combine three lexica (our lexicon and two verb lexica) into a single vector space, making connotations easier to use as model input and providing a single representation method for the connotations of any word.

We design a novel multi-task learning model that jointly predicts all of the connotation labels for a word $w$, from a learned representation $v_w$. Each task is to predict the label for exactly one connotation aspect: 
% one of 
the 6 aspects in \S \ref{sec:labels} for nouns and adjectives 
and
% one of 
the
11 aspects in 
CFs+ (connotation frames and their extension to power and agency) for verbs 
\citep{rashkin-etal-2016-connotation,Sap2017ConnotationFO}.

To learn a representation for $w$ we encode dictionary definitions of the word $w$ and words related to $w$ (e.g., synonyms, hypernyms) in a single vector, which we then use to predict connotation labels. We use definitions and related words since linguists have argued that definitions and related words convey a word's meaning~\cite{GauralnikDict}.

\begin{figure}[t]
    \centering
    \includegraphics[width=.3\textwidth]{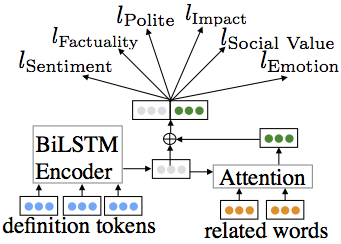}
    \caption{Connotation embedding modeling in \S \ref{sec:connmethod}.} 
    \label{fig:arch}
    \vspace{-10pt}
\end{figure}

Let $w$ be a word with part of speech $t$. The input to the connotation encoding model is then: (1) a set of dictionary definitions $D_{w^t}$ and (2) a set of words related to $w^t$, $R_{w^t}$. We use multiple definitions to capture multiple senses of $w^t$. To emphasize more prevalent senses of $w^t$, we use similar repeated definitions for the same sense, collected from multiple sources. From $D_{w^t}$ and $R_{w^t}$, the encoder produces a connotation feature embedding $v_{w^t} \in \mathbb{R}^d$ of dimension $d = 300$. Then we use $v_{w^t}$ to predict the label $\ell_a$ for 
% each 
connotation aspect $a$ (see Figure~\ref{fig:arch}).

\subsubsection{Encoding Models}
\label{sec:enc}
For a word $w^t$, the input to our encoder is $d_{w^t} = [d_{w^t}^1; d_{w^t}^2; ...; d_{w^t}^N] \in \mathbb{R}^{Nd_{in}}$, the sequence of fixed pre-trained token embeddings for concatenated definitions in $D_{w^t}$.
Then we take as our embedding the normalized final hidden state from a BiLSTM, a standard architecture for text encoding: 
$v_{w^t} = \frac{h_{w^t}}{\|h_{w^t}\|}$, where $h_{w^t} = \text{BiLSTM}(d_{w^t})$ and
$h_{w^t}  \in \mathbb{R}^{2H}$ is the concatenation of the last forward and backward hidden states (model \textbf{\texttt{CE}}).

As a variation of our model, we apply scaled dot-product attention~\cite{VaswaniAttention} over the related words $R_{w^t}$, using $h_{w^t}$ as the attention query, to obtain $v_{w^t}$. Then we add the result to $h_{w^t}$ before normalizing
(model \textbf{\texttt{CE+R}}).

\subsubsection{Label Classifier}
For each connotation aspect, we train a separate linear layer plus softmax with the input $[v_{w^t}; e_{w^t}]$, where $e_{w^t}$ is the pre-trained embedding for $w^t$. For the non-emotion aspects, the layer has three target classes $\{-1, 0, 1\}$ for most aspects (four classes for the `power' and `agency' verb aspects) and we predict the label with highest output probability. For emotions, we do multi-label classification by thresholding the output probabilities for each emotion dimension with a fixed $\theta \in \R$. 
We include $e_{w^t}$ in the predictor input to encourage $v_{w^t}$ to model connotation information that is complementary to the information present in pre-trained embeddings. 

\subsubsection{Learning}
\label{sec:learning}
For each non-emotion connotation aspect $a$ (e.g., \textit{Impact}) we calculate the weighted cross-entropy loss $\mathcal{L}^a$. For \textit{Emo} we calculate the one-versus-all cross-entropy loss on each of the eight emotions, $\mathcal{L}_i^{Emo}$ 
for $1 \leq i \leq 8$,
and 
their sum is
% sum them to obtain 
$\mathcal{L}^{Emo}$. 

In our multi-task joint learning framework \textbf{\texttt{(J)}}, we minimize the weighted sum of $\mathcal{L}^a$ across all connotation aspects. 
% (models ).
We also experiment with 
% learning each connotation label classifier individually,
training a separate encoding model for each connotation aspect $a$ that minimizes $\mathcal{L}^a$ \textbf{\texttt{(S)}}.
% (models ). 

\subsubsection{Baselines and Models}
For each baseline, we implement one classifier per connotation aspect, or, for \textit{Emo}, one classifier for each emotion.
Following Rashkin et al.~\shortcite{rashkin-etal-2016-connotation} we implement a Logistic Regression classifier trained on the 300-dimensional pre-trained word embedding for $w$ using the standard L-BFGS optimization algorithm and sample re-weighting (\texttt{LR}). 
We also implement a majority class baseline (\texttt{Maj}).

We present three variations of our model: (i) trained jointly for all parts of speech and all connotation aspects (\texttt{CE(J)}), (ii)  trained on each aspect individually with related word attention (\texttt{CE+R(S)}), and (iii) trained jointly on all parts of speech and all connotation aspects with related word attention (\texttt{CE+R(J)}).

%%%%%%%%%%%%%%%%%%%%%%%%%%
% connotation prediction
%%%%%%%%%%%%%%%%%%%%%%%%
\subsection{Connotation Prediction}
\label{sec:connpred}
\subsubsection{Data and Parameters}
For nouns and adjectives, we train using the aspects described in \S \ref{sec:lex} (6 aspects). For verbs, we train on 9 aspects\footnote{perspective of the writer on the agent/theme, perspective of the agent on the theme, effect on/value/ state of entities} from Rashkin et al.~\shortcite{rashkin-etal-2016-connotation} as well as `power' and `agency' from Sap et al.~\shortcite{Sap2017ConnotationFO} (11 aspects total). We split our connotation lexicon (\S \ref{sec:lex}) into train (60\%), development (20\%) and test (20\%). 
For the verb CFs+, we preserve the originally published data splits where possible. We move words only to ensure that all parts of speech for a word are in the same split 
% (e.g., `evil' (noun) and `evil' (adj)
(e.g., `evil' both as a noun and adj is in the dev set).
% are both in the development set). 

We collect dictionary definitions and related words from all seven dictionaries available on the Wordnik API\footnote{\url{https://www.wordnik.com/}}. These are extracted for each word and part-of-speech pair. We preprocess definitions by removing stopwords, punctuation, and the word itself. We 
use
% train our models 
%using 
pre-trained Concept-Net numberbatch embeddings~\cite{numberbatch}. 
% , with $d_{in}=300$

\begin{table}[t]
\centering
\begin{tabular}{|l|lllll|}
\hline
 & \textbf{Maj} & \textbf{LR} & \begin{tabular}[c]{@{}l@{}}\textbf{CE+R}\\ \textbf{(S)}\end{tabular} & \begin{tabular}[c]{@{}l@{}}\textbf{CE}\\ \textbf{(J)}\end{tabular}  & \begin{tabular}[c]{@{}l@{}}\textbf{CE+R}\\ \textbf{(J)}\end{tabular} \\ \hline
\textit{N/Aj Avg}
& .304 & .594 & .589 & .597 & .597  \\ 
\textit{Verb Avg}
& .222 & .553 & .489 & .521  & .520 \\ 
\hline
\textbf{\textit{All Avg}}
& .251 & .568 & .524 & .548 & .547 \\ \hline
\end{tabular}
\caption{Macro-averaged F1 for connotation prediction on the test set, averaged across aspects. N/Aj indicates noun and adjective.}
\label{tab:results}
\end{table}
\subsubsection{Results}
\textbf{Label Prediction} \\
We present results on the connotation prediction task to check
the quality of our representation learning system.
% that our representation learning system is working. 
Given dictionary definitions and related words, we predict
the labels from our lexicon (\S \ref{sec:lex}) and CFs+
(see Table \ref{tab:results}).

First, we observe that joint learning (models \texttt{(J)}) improves over training representations individually (\texttt{CE+R(S)}). We hypothesize that 
joint learning provides regularization across all aspects. Second, we compare joint learning with (\texttt{CE+R(J)}) and without (\texttt{CE(J}) related words to the strong \texttt{LR} baseline. 
We find that the model with related words (\texttt{CE+R(J)}) is statistically indistinguishable from the baseline\footnote{We use an approximate randomization test}
(for $p \leq 0.05$).
In contrast, our model without related words (\texttt{CE(J)}) is significantly worse than the \texttt{LR} baseline for one aspect (see Appendix \ref{app:connmodel} for aspect-level results).
Thus we conclude that related words are beneficial for learning connotations. \\ \indent
Overall, our approach provides a single unified feature representation for the lexical connotations of all parts of speech, without any loss in label prediction performance.
Specifically, our best representation learning model (\texttt{CE+R(J)}) has comparable label prediction performance to a strong baseline (\texttt{LR}),
a baseline that does not learn any kind of representation. 
We use \texttt{CE+R(J)} to generate connotation embeddings 
that we use in all further evaluation.\\

\begin{table*}[ht]
\begin{tabular}{lllll}
\hline
\textbf{Aspect} & \textbf{Word} & \textbf{Conn Only} & \textbf{Both} & \textbf{Word Only} \\ \hline

% \multirow{2}{*}{\begin{tabular}[c]{@{}l@{}}\textit{Social}\\ \textit{Value}\end{tabular}} 
\textit{Social Value}& ability (+) & \begin{tabular}[c]{@{}l@{}}service, imagination, \\ worth, practical\end{tabular} & NONE & \begin{tabular}[c]{@{}l@{}}lack, inability, \\ enough, difficult\end{tabular} \\ \hline
%  & abject (-) & \begin{tabular}[c]{@{}l@{}}disgusting, ugly, \\ filthy, missing,\end{tabular} & \begin{tabular}[c]{@{}l@{}}miserable, woeful, \\ atrocious, gastly, \\ idelness\end{tabular} & \begin{tabular}[c]{@{}l@{}}pathetic, earnestness, \\ subservience, failure,\\ futility\end{tabular} \\ \hline
\multirow{2}{*}{\textit{Polite}} & slug (-) & \begin{tabular}[c]{@{}l@{}}bang, shove, \\ murder, scum\end{tabular} & NONE  & \begin{tabular}[c]{@{}l@{}}quote, bug, exception, \\ reference\end{tabular} \\ \hline
%  & leadership (+) & \begin{tabular}[c]{@{}l@{}}museum, account, \\ parliament, throne\end{tabular} & constitution, legislation & \begin{tabular}[c]{@{}l@{}}addition, outset, \\ authority figure, struggle\end{tabular} \\ \hline
% \multirow{2}{*}{\textit{Impact}} & fraud (-) & \begin{tabular}[c]{@{}l@{}}strain, deter, crime, \\ robbery, subversion\end{tabular} & abuse & \begin{tabular}[c]{@{}l@{}}evasion, failure, related, \\ bogus, result, suspicion\end{tabular} \\
\multirow{2}{*}{\textit{Impact}} & merry (+) & \begin{tabular}[c]{@{}l@{}}glee, exhilaration, \\ pretty, prosperous\end{tabular} & \begin{tabular}[c]{@{}l@{}}cheery, genial, joyful, \\ fun, merriment, jolly\\ joy, delightful, cheerful\end{tabular} & \begin{tabular}[c]{@{}l@{}}amiable, wonderful,  \\ crazy, wives\end{tabular} \\ \hline
\end{tabular}
\caption{Examples of nearest neighbors in the connotation embedding space (Conn Only), the pre-trained word embedding space (Word Only) and all 
top 50 nearest neighbors in both spaces (Both). NONE indicats no overlap.
% the words in the nearest neighbors in both spaces (Both). NONE indicates no words in the top 50 are nearest neighbors in both spaces.
}\label{tab:knnex}
\end{table*}
\noindent
\textbf{Observations}\\
Our connotation representation learning model presents several advantages. Since the model uses dictionary definitions,
% For example, it can generate a representation for a word in a zero-shot manner from only a few dictionary definitions, rather than the thousands of examples of contextual use required by standard word-embedding methods.
% For example, 
we can generate representations for slang words (e.g., ``gucci'' meaning ``really good''), where knowledge-base entries (e.g., in Concept-Net) do not capture the slang meaning. For example, in our connotation embedding space, the nearest neighbors of ``gucci'' include words related to the slang connotations (e.g., ``beneficial'' -- positive impact, not factual), whereas neighbors in a pre-trained word embedding space are specific to the fashion meaning and connotations (e.g., ``buy'', ``italy'', ``textile''). Along with slang, our model can also generate representations for new or rare words (e.g., ``merchantile'') that don't have a pre-trained word representation.
% We demonstrate further uses of our connotation representation $v_{w^t}$ in intrinsic (\S \ref{sec:inteval}) and extrinsic (\S \ref{sec:exteval}) evaluation.

%%%%%%%%%%%%%%%
%% Experiments
%%%%%%%%%%%%%%%%
\section{Experiments}
\label{sec:exp}

%%%%%%%%%%%%%%%%%%%%%%%%
% intrinsic evaluation %
%%%%%%%%%%%%%%%%%%%%%%%%

\subsection{Intrinsic Evaluation}
\label{sec:inteval}
To evaluate the connotation embedding space, we look at the
$50$ nearest neighbors, by Euclidean distance, of every word in our training and development sets. 
We find that neighbors in the connotation embedding space
are more closely related based on the connotation label than in the pre-trained embedding space.

Looking at example nearest neighbors (Table~\ref{tab:knnex}) we see that nearest neighbors in the pre-trained embedding space include antonyms (e.g., ``inability'' is close to ``ability'') and topically related words (e.g., ``merry'' is close to ``wives''), while in the connotation space, neighbors often share connotation labels even though they may be
%KM - I added a comma after the while clause. See if you are OK with that.
% includes more examples with the same connotation label for some aspect but are 
topically or denotatively unrelated. For example, ``slug'' (noun) is close to many \textit{impolite} but otherwise unrelated words (e.g., ``shove'', ``murder'', ``scum'') in the connotation embedding space while in the pre-trained space ``slug'' is close to topically related (e.g., ``bug'') but \textit{polite} words. Therefore, we can see that 
words with similar connotations are placed closer together than in the pre-trained semantic space.
% the connotation feature embeddings place words with similar connotations closer together, reshaping the pre-trained semantic space. 

To quantify the semantic differences, we measure neighbor-cluster connotation label purity. Specifically, for each connotation aspect $a$ 
(e.g., \textit{Social Value}) and each non-neutral label $c$ (e.g., valuable (+)), 
we calculate $r_c^{a(C)}$: the average ratio of words with label $c$ to label $-c$ in the set of nearest neighbors of all words with label $c$ for aspect $a$.
We compare it against the same ratio for the nearest neighbors selected using the same pre-trained word embeddings as in \S \ref{sec:connpred}, denoted $r_c^{a(P)}$.

\begin{table}[t]
\centering
\begin{tabular}{llrrr}
\hline
& $\boldsymbol{c}$ & \textbf{\textit{Social Val}}
% \begin{tabular}[c]{@{}l@{}}\textit{Social}\\ \textit{Value}\end{tabular} 
& \textbf{\textit{Polite}} & \textbf{\textit{Impact}}\\ \hline
\multirow{2}{*}{$\boldsymbol{r_c^{a(C)}}$} & + & 21.27 & 2640.14 & 47.49\\
& - & 5.88 & 50.00 & 33.33\\ \hline
\multirow{2}{*}{$\boldsymbol{r_c^{a(P)}}$} & + & 4.70 & 43.71 & 4.73\\
& - & 2.38 & 0.54 & 8.33\\ \hline

%  \textbf{Aspect } $\boldsymbol{a}$ & $\boldsymbol{c}$ & $\boldsymbol{r_c^{a(C)}}$ & $\boldsymbol{r_c^{a(P)}}$\\ \hline
% \multirow{2}{*}{\begin{tabular}[c]{@{}l@{}}\textit{Social}\\ \textit{Value}\end{tabular}} & + & 21.27 & 4.70\\ 
%   & $-$ & 5.88 & 2.38 \\ \hline
% \multirow{2}{*}{\textit{Polite}} & + & 2640.14 & 43.71\\ 
%   & $-$ & 50.00 & 0.54\\ \hline
% \multirow{2}{*}{\textit{Impact}} & + & 47.49 & 4.73\\ 
% & $-$ & 33.33 & 8.33 \\ \hline
% \multirow{2}{*}{\textit{Fact}} & + & 0.84 & 0.37 \\ 
%   & $-$ & 4.00 & 9.09 \\ \hline
% \multirow{2}{*}{\textit{Sent}} & + & 4.73 & 1.33 \\ 
%  & $-$ & 1.41 & 1.11 \\ \hline
% \multirow{2}{*}{
% \textit{Emo Avg}
% } & $\geq1$ & 0.28 & 0.27  \\ 
% & 0 & 33.33 & 20.00 \\ \hline
\end{tabular}
\caption{Select cluster connotation purity ratios. 
% \textit{Emo} is based on no emotions $(0)$ or any number $(\geq 1)$.
}
\label{tab:knn}
\end{table}
We find that across connotation aspects, these ratios are higher for the learned connotation embeddings, compared to pre-trained embeddings. For example, $r_+^{\textit{Social Val}(C)} = 21.27$ but $r_+^{\textit{Social Val}(P)} = 4.70$
(see Table \ref{tab:knn}).
This shows the connotation 
embeddings 
reshape the pre-trained semantic space.
% cluster words based more on connotation labels than pre-trained embeddings.

%%%%%%%%%%%%%%%%%%%%%%%%
% extrinsic evaluation %
%%%%%%%%%%%%%%%%%%%%%%%%
\subsection{Extrinsic Evaluation}
\label{sec:exteval}
%KM Check following cut
We further evaluate our connotation embeddings using the stance detection task, hypothesizing  they will lead to improvement.
%Stance is often expressed through subtle language, and we hypothesize that connotations can improve stance detection. 
Given a text on a topic (e.g., ``gun control''), the task is to predict the stance (pro/con/neutral) towards the topic
%: whether the text is pro/con/neutral towards the topic
% supports the topic, is against the topic, or is neutral 
(see Figure~\ref{fig:intro}).

\subsubsection{Methods and Experiments}
\begin{figure}[t]
    \centering
    \includegraphics[width=.39\textwidth]{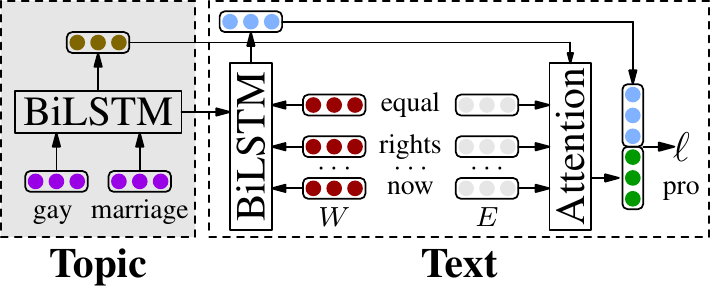}
    \vspace{-10pt}
    \caption{Stance models \texttt{BiC+}$E$; $E \in \{C, W, R\}$.}
    \vspace{-10pt}
    \label{fig:stancearch}
\end{figure}

\noindent
\textbf{Models}\\ As a baseline architecture, we implement the bidirectional 
conditional encoding model~\cite{Augenstein2016StanceDW}. 
This model encodes a text as $h_T$ with a BiLSTM, conditioned on a separate topic encoding $h_P$, and predicts stance from $h_T$ (\texttt{BiC}). 
We include connotation embeddings through scaled dot-product attention over the noun, adjective, and verb embeddings from the text, with $h_P$ as the query
(see Figure~\ref{fig:stancearch}).
We experiment with three types of embeddings in the attention: pre-trained word embeddings (\texttt{BiC+W}), our connotation embeddings (\texttt{BiC+C}), and randomly initialized embeddings (\texttt{BiC+R}), as a baseline to measure the importance of attention. We also implement a Bag-of-Word-Vectors baseline (\texttt{BoWV}), encoding the text and topic as separate BoW vectors and passing their concatenation to a Logistic Regression classifier.\\

\noindent
\textbf{Data and Parameters}\\ We use the Internet Argument Corpus~\cite{Abbott2016InternetAC}: ${\sim}59k$ posts
% on $16$ topics 
from online debate forums. 
% in our experiments 
% (see Table~\ref{tab:stancedata}).
Of the $16$ total topics, four are large (\textit{L}, with $>7k$ ex each), five are medium (\textit{M}, with ${\sim}2k$ ex each), and seven are small (\textit{S}, with $30$-$300$ ex each). 

Since not every text will take a position on every topic, 
we automatically generate `neutral' examples for the data. To do this, we sample a pro/con example and then assign it a new (different) topic, randomly sampled from the original topic distribution. 
% we augment the data with examples for the `neutral' class where each example is assigned a new topic, different from the original, that we randomly sample from the original topic distribution. 
We split the data into train, development, and test such that no posts by one author are in multiple splits and preprocess the data by removing stopwords and punctuation and lowercasing.

Stance is topic-dependent and as a result, models require numerous training examples for each individual topic. However, many examples are not always available for every topic. Since there are hundreds of thousands of potential topics, the vast majority of which will have very few examples, our goal is to build models that exhibit strong performance across all topics, regardless of size. 
Therefore, we
experiment with three data scenarios: (i) training and evaluating using all the data (\texttt{All Data}), (ii) truncating each topic in training to \textit{M} size (at most $2k$ examples) and evaluating using all data (\texttt{Trunc Train}), and (iii) truncating each topic to \textit{M} size in training and in evaluation (\texttt{Trunc All}), so that topics have the same frequency for both training and evaluation.

\subsubsection{Results}
\begin{figure}[t]
    \vspace{-20pt}
    \centering
    \includegraphics[width=.49\textwidth]{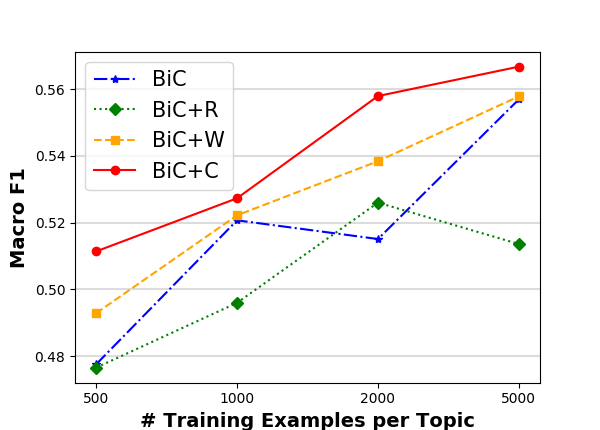}
    \caption{Stance F1 on the test set as number of training examples per topic varies.}
    \label{fig:datasamples}
    \vspace{-8pt}
\end{figure}

\begin{table}[t]
\centering
\begin{tabular}{llll}
\hline
& All Data & Trunc Train & Trunc All \\ \hline
\textbf{\texttt{BoWV}} & .3587 & .3473 & .3613\\
\textbf{\texttt{BiC}} & .5677 & .5151 & .5244\\
\textbf{\texttt{BiC+R}} & .5282 & .5260 & .5128\\
\textbf{\texttt{BiC+W}} & .5650 & .5384 & .5421\\
\textbf{\texttt{BiC+C}} & .5613 & .5579$^*$ & .5562$^*$\\ \hline

%  \textbf{Data Size}
%  & \textbf{\texttt{BoWV}} & \textbf{\texttt{BiC}}
%  & \begin{tabular}[c]{@{}l@{}}\textbf{\texttt{BiC}}\\\textbf{\texttt{ +R}}\end{tabular} & \begin{tabular}[c]{@{}l@{}}\textbf{\texttt{BiC}}\\ \textbf{\texttt{ +W}}\end{tabular} & \begin{tabular}[c]{@{}l@{}}\textbf{\texttt{BiC}}\\\textbf{\texttt{ +C}}\end{tabular} \\ \hline
% All Data
% & .3587 & .5677 & .5282 & .5650 & .5613 \\ %\hline
% Trunc Train
% & .3473 & .5151 & .5260 & .5384 & .5579$^*$ \\ %\hline
% Trunc All
% & .3613 & .5244 & .5128 & .5421 & .5562$^*$ \\ \hline
\end{tabular}
\caption{Stance detection macro-averaged F1 on the test set. 
$^*$ indicates significance ($p < 0.01$) between \texttt{BiC+C} and \texttt{BiC+W}.}\label{tab:stance}
\end{table}
\indent
We find that when using all of the training data, the pre-trained embeddings and our connotation embeddings perform comparably (significance level $p=0.3$). 
Note that both the connotation and pre-trained embeddings outperform the random embeddings in all scenarios, showing that the architecture difference is not the only reason for improvement when adding embeddings. 
We find that in both scenarios where data is limited per topic (\texttt{Trunc Train} and \texttt{Trunc All}), the connotation embeddings improve significantly over the pre-trained word embeddings. In fact, the same trend is visible across varying numbers of training examples (see Figure~\ref{fig:datasamples}).
Our results demonstrate that the connotation information is useful for detecting stance when data is limited.

We find further evidence that the connotation embeddings (\texttt{BiC+C}) make the model robust to loss
of training data when we look at the results on the individual topic level. Namely, in setting \texttt{Trunc Train}, \texttt{BiC+C} has a significant improvement (with $p <0.05$) over \texttt{BiC+W} on six topics, including
four of the \textit{M} and truncated \textit{L} topics. In fact, for the four \textit{M}/\textit{L} topics, the average per-topic decrease in F1 for \texttt{BiC+C} is $1/4$ that of \texttt{BiC+W}. These per-topic results further highlight the robustness of \texttt{BiC+C} when training data is restricted. \\ \indent
We conclude that connotation embeddings improve stance performance when training data is limited, suggesting they can be used in future work that generalizes stance models to topics with no training data (i.e. most topics).

\section{Conclusion}
We create a new lexicon with six new connotation aspects for nouns and adjectives that aligns well with human judgments. We also show that the lexicon confirms hypotheses about semantic divergences between synonyms. We then use our lexicon to train a unified connotation representation for words from all parts of speech, yielding an embedding space that captures more connotative information than pre-trained word embeddings.  

We evaluate our connotation representations on stance detection. Since the stance detection tasks encountered in real life concern a very large number of topics, zero-shot and few-shot stance detection are important subtasks.
We show that models using our connotation representations are well suited for few-shot stance detection and
may also
generalize well to zero-shot settings. 

In future work, we plan to explore the relationships between connotations, context, and word sense, as well as adapting our methods to learn multi-lingual connotation representations that accurately capture cultural and linguistic variations. 

\section*{Acknowledgements}
We thank the Columbia NLP group and the anonymous reviewers for their comments. This work is based on research sponsored by DARPA under agreement number FA8750-18-
2-0014. The U.S. Government is authorized to reproduce and distribute reprints for Governmental purposes notwithstanding any copyright notation thereon. The views and conclusions contained herein are those of the authors and should not be interpreted as necessarily representing the official policies or endorsements, either expressed or implied, of DARPA or the U.S. Government.

% include your own bib file like this:
\bibliography{anthology,eacl2021,ref}
\bibliographystyle{acl_natbib}

\newpage
\appendix

\section{Overview}
The data and software are provided as supplementary material her: \url{https://github.com/connotationembeddingteam/connotation-embedding}.

\begin{table*}[ht]
\centering
\begin{tabular}{ll}
\hline
\textbf{Aspect} & \textbf{General Inquirer Categories} \\
\hline
\textit{Social Value} & \begin{tabular}[c]{@{}l@{}}PowGain, PowLoss, PowEnds, PowCon, PowCoop, PowAuPt, PowPt, PowAuth,\\ PowOth,
RcEthic, RcRelig, RcGain, RcEnds, RcLoss, Virtue, Vice, WltPt\\ WltTran, WltOth, Food, Object, Doctrin, Academ, Work, NatrObj, Vehicle, Econ@,\\ Goal, EnlPt, EnlOth, EnlLoss, SklPt, SklAsth, SklOth, Exprsv, Legal, COLL, Means\\ MeansLw, Fail, Solve, EndsLw, Try, WlbPhys, WlbGain, WlbPt, WlbLoss, WlbPsyc,\\ Quality, SocRel \end{tabular}\\ \hline
\textit{Politeness} & \begin{tabular}[c]{@{}l@{}} RspGain, RspLoss, RspOth, AffGain, AffLoss, AffOth, WlbPt, SklPt, EnlPt, Relig,\\ WltPt, Polit\@, HU, Milit, Legal, Academ, Doctrin \end{tabular} \\ \hline
\textit{Impact} & \begin{tabular}[c]{@{}l@{}}PosAff, Pleasur, Pain, NegAff, Anomie, NotLw, Vice, Virtue, RcGain, RcLoss,\\ RspLoss, RcEthic, RspOth, WlbPysc, RcEnds, EnlOth, WlbGain, RspGain, EnlGain,\\ EnlEnds, EnlPt, WlbLoss, WlbPt, EnlLoss, SklOth, WlbPhys, Try, Goal, Work \end{tabular}\\  \hline
\textit{Factuality} & $v = \begin{cases}
    -1 & \mbox{if } x \leq -0.25 \\
    1 & \mbox{if } x \geq 0.25\\
    0 & \mbox{otherwise}\\
\end{cases}$
where $x$ is the Imagery score normalized to $[-1, 1]$. \\ \hline
\textit{Sentiment} & $v = \begin{cases}
    -1 & \mbox{if } x \leq -0.25 \\
    1 & \mbox{if } x \geq 0.25\\
    0 & \mbox{otherwise}\\
\end{cases}$
where $x$ is the sentiment score normalized to $[-1, 1]$. \\ \hline
\end{tabular}
\caption{Categories from the Harvard General Inquirer used in distant labeling connotations.}\label{tab:hgicats}
\end{table*}

\section{Connotation Labeling}
\label{app:connlabel}
We construct labels for our connotation lexicon (\S \ref{sec:lex}) using categories from the following existing resources: \textit{HGI} -- the Harvard General Inquirer~\citep{Stone1963ACA}, \textit{DAL} -- the revised Dictionary of Affect in Language~\citep{Whissell2009UsingTR}, \textit{CWN} -- Connotation WordNet~\citep{kang-etal-2014-connotationwordnet}, and \textit{NRCEmoLex} -- the NRC Emotion Lexicon~\citep{Mohammad13}. 

The HGI consists of $183$ psycho-sociological categories. Each lexical entry (${\sim}11k$ total) is tagged with a non-zero number of categories. Different senses (noted through brief definitions) and parts-of-speech for the same word have separate entries. The available categories include valence (i.e., positive and negative), words related to a particular entity or social structure (e.g., institutions, communication), and value judgements (e.g., concern with respect).

The DAL consists of ${\sim}8k$ words with scores for 3 categories: pleasantness, activation, and imagery. Word entries include inflection but do not explicitly mark part-of-speech. CWN is a lexicon of connotation polarity scores (ranging from $0$ to $1$) for ${\sim}180k$ words, explicitly marked for part of speech. Finally, NRCEmoLex consists of word entries marked for any number the eight Plutchik emotions (anticipation, joy, trust, fear, surprise, sadness, disgust, anger) as well as positive and negative sentiment. Two versions of the lexicon are available: with and without sense level distinctions. Neither version includes explicit information on part-of-speech, and so we infer part-of-speech using the words provided to distinguish different senses. 

We provide the complete distant labeling rules for each of the connotation aspects in Table~\ref{tab:hgicats} (see \url{http://www.wjh.harvard.edu/~inquirer/homecat.htm} for complete information on abbreviations). Within each connotation aspect, we determine the connotation polarity using the additional categories: \textit{Positiv, Negativ, Strong, Weak, Hostile, Submit, Active} and \textit{Power}.

\section{Analysis of the Connotation Lexicon}
In this section, we provide further analysis of the connotation lexicon as exemplification of its content and properties.

\subsection{Human Evaluation}
\begin{figure*}[ht]
    \vspace{-100pt}
    \centering
    \includegraphics[width=0.98\textwidth]{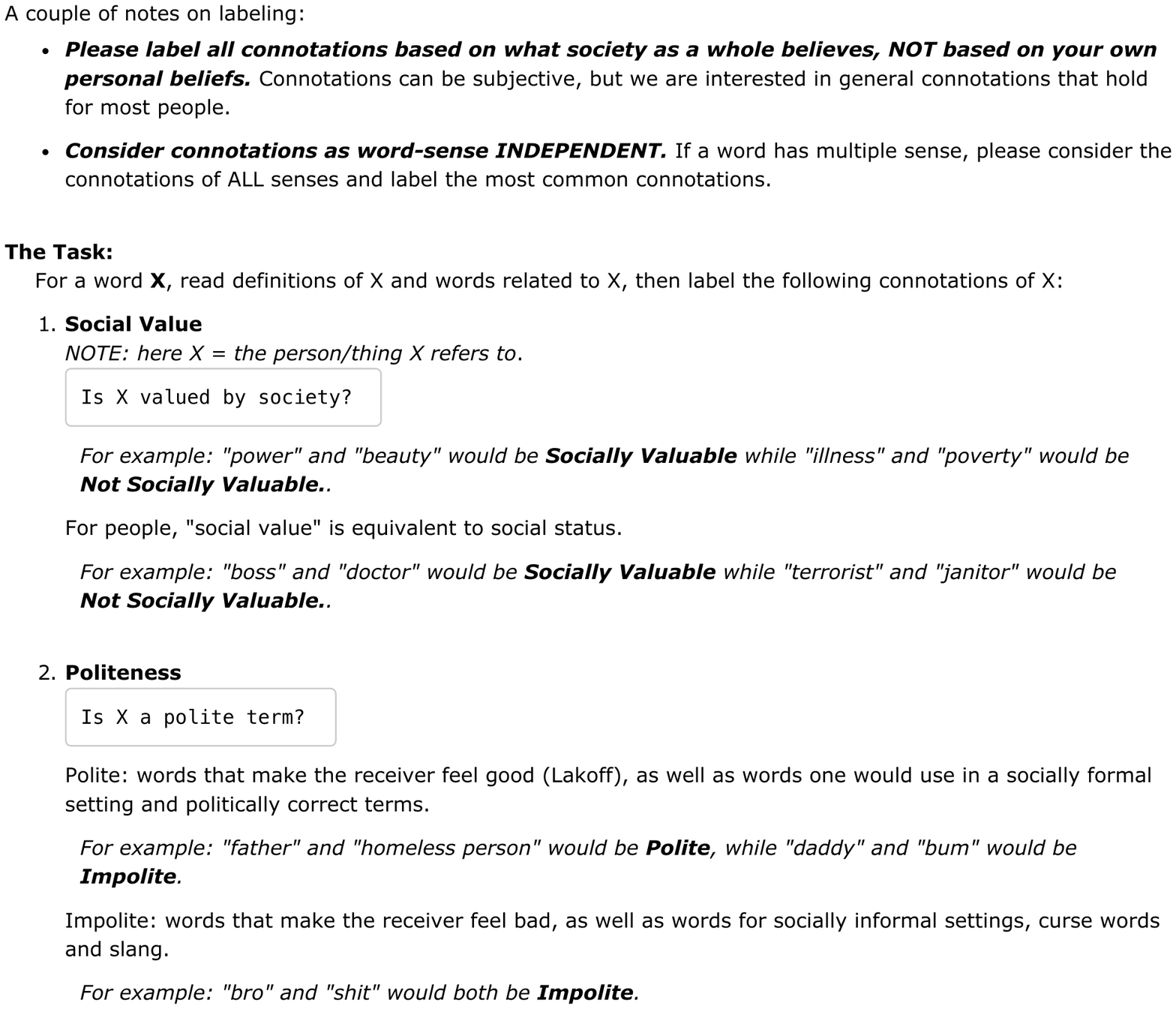}
    \vspace{-130pt}
    \caption{First part of annotator instructions for connotation labeling in \S \ref{sec:conneval}.} 
    \label{fig:appanninst1}
\end{figure*}

\begin{figure*}[ht]
\vspace{-230pt}
    \centering
    \includegraphics[width=0.98\textwidth]{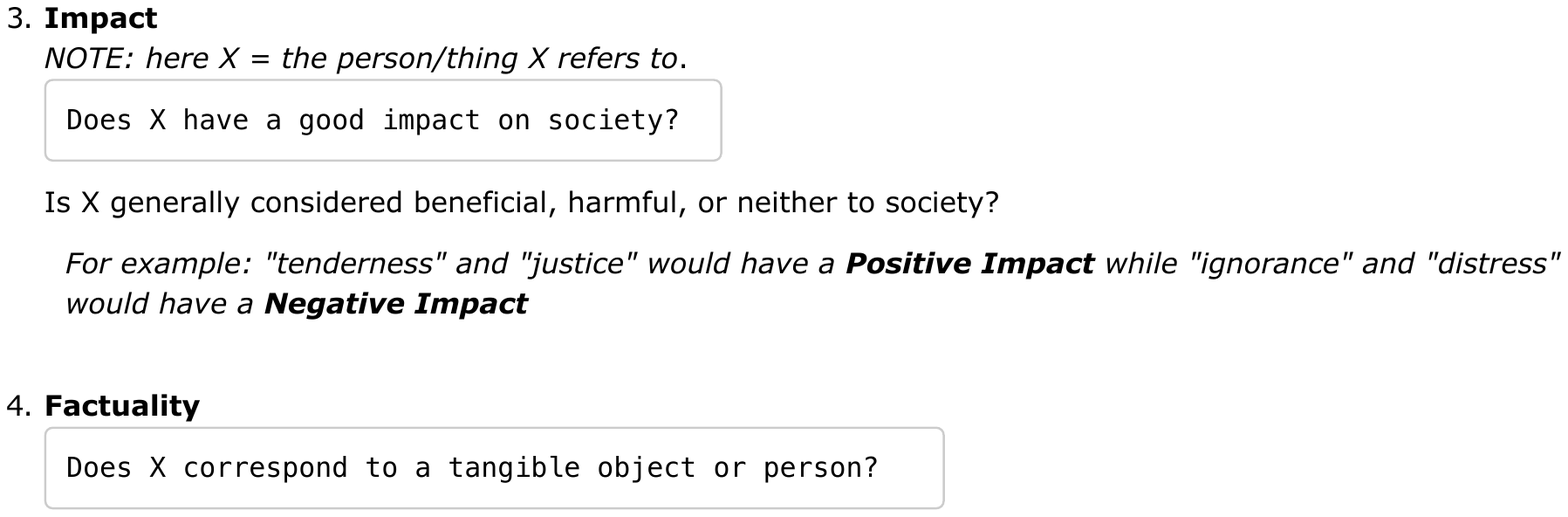}
    \vspace{-230pt}
    \caption{Second part of annotator instructions for connotation labeling in \S \ref{sec:conneval}.} 
    \label{fig:appanninst2}
\end{figure*}

We show the instructions provided to annotators for the manual labeling of samples from the connotation lexicon in \S \ref{sec:conneval} (see Figures \ref{fig:appanninst1} and \ref{fig:appanninst2}).

We include Cohen's $kappa$ score for agreement between the lexicon and human annotators for individual connotation aspects in Table~\ref{tab:appcohenkappas}. 
\begin{table}[t]
\centering
\begin{tabular}{lc}
\hline
\textbf{Aspect} & \begin{tabular}[c]{@{}l@{}}\textbf{Avg}\\ \textbf{Cohen's $\kappa$}\end{tabular} \\ \hline
\multicolumn{1}{l|}{\begin{tabular}[c]{@{}l@{}}\textit{Social} \\ \textit{Value}\end{tabular}} & 0.526\\ %\hline
\multicolumn{1}{l|}{\textit{Politeness}} & 0.186\\ %\hline
\multicolumn{1}{l|}{\textit{Impact}} & 0.595\\ %\hline
\multicolumn{1}{l|}{\textit{Factuality}} & 0.164\\ \hline
\multicolumn{1}{l|}{\textbf{Average}} & 0.368\\ \hline
\end{tabular}
\caption{Cohen's $\kappa$ for agreement between the connotation lexicon and human annotators. }\label{tab:appcohenkappas}
\end{table}

\subsection{Gender Bias Analysis}
Connotations have  been used to study gender bias in movie scripts \citep{Sap2017ConnotationFO} and online media \citep{Field2019ContextualAA}. Here we use our connotation lexicon to analyze gender bias in two new domains: celebrity news (Celebrity) and student reviews of computer science professors (Professors). 

We use existing datasets for these domains and the accompanying methodology of \citet{Chang2019AutomaticallyIG} to infer word-level gender associations. Then, for the gender-associated words that are in our lexicon, for each connotation aspect and 
domain, 
we examine the percentage of positive and negative polarity words and find that these quantify known trends in gender-biased portrayals.

In the Celebrity domain, \textit{Factuality} highlights the tendency of news media to focus on physical characteristics of female celebrities~\citep{Selby}. More words with 
positive \textit{Factuality} polarity (tangible concepts and attributes) are associated with women and more words with negative polarity (abstract concepts and attributes) are associated with men. For example, women are described as ``beautiful'' and ``slim'', while men are described as ``political'' and ``presidential''. In fact, even many of the not tangible female-associated words still align with physical attributes (e.g., ``chic''), further emphasizing the biased portrayal. 

We also find that in the Professors domain, patterns in \textit{Social Value} and \textit{Impact} agree with the observations of \citet{Chang2019AutomaticallyIG} and with social science literature that finds male teachers are praised more than female teachers for being experts (both socially valuable and positively impacting society). For example, men are associated with positive \textit{Social Value} (socially valuable) words such as ``knowledge'' and ``experience'', while women are relatively less often associated with the same type of words.

Finally, in the Celebrity domain we find our lexicon reflects the coverage in the media of recent sexual harassment allegations against male celebrities.
Namely, women are associated with more positive \textit{Social Value} words and men are associated with many more negative \textit{Social Value} words (see Table \ref{tab:genderex}). 
Overall, our results quantitatively validate previous observations and known patterns of gender bias. 

\begin{table}[t]
\begin{tabular}{llll}
\centering
\textbf{Aspect} & \textbf{Pol} & \textbf{F:M} & \textbf{Examples} \\ \hline
\multirow{4}{*}{\textit{Fact}} & \multirow{2}{*}{+} & \multirow{2}{*}{54:27} & 
% \rowcolor{lightgray} 
F: beautiful, slim \\ 
%\cline{4-4} 
 &  &  &  M: actor, film \\ 
%  \cline{2-4} 
 &  \multirow{2}{*}{$-$} & \multirow{2}{*}{26:39} & 
% \rowcolor{lightgray} 
F: style, chic\\ 
 %\cline{4-4} 
 &  &  & M: apology, political \\ \hline
\multirow{4}{*}{\begin{tabular}[c]{@{}l@{}}\textit{Social}\\ \textit{Value}\end{tabular}} & \multirow{2}{*}{+} & \multirow{2}{*}{39:27} & 
% \rowcolor{lightgray} 
F:  beauty, body, gold \\ 
%\cline{4-4} 
 &  &  & M: attempt, evidence \\ 
%  \cline{2-4} 
 & \multirow{2}{*}{$-$} & \multirow{2}{*}{3:24} &
% \rowcolor{lightgray} 
F: blue, dancer, party \\ 
 %\cline{4-4} 
 &  &  & M: abuse, allegation \\ \hline
\end{tabular}
\caption{Gender bias examples in the Celebrity news domain. Pol is Polarity (tangible(+) vs. not tangible($-$) and valuable (+) vs. not valuable ($-$)). F:M shows percent of female-associated words to percent of male-associated.}\label{tab:genderex}
\end{table}

\section{Connotation Modeling}
\label{app:connmodel}
\subsection{Data} We use dictionary definitions extracted from all available dictionaries on the Wordnik API: American Heritage Dictionary, CMU Pronouncing Dictionary, Macmillan Dictionary, Wiktionary, Webster's Dictionary, and WordNet. For labels, we use six aspects (see \S \ref{sec:lex}): \textit{Social Value}, \textit{Politeness}, \textit{Impact}, \textit{Factuality}, \textit{Sentiment}, and \textit{Emotional Association}. For verbs we use 11 aspects: perspective of the writer on the theme \textit{P(wt)} and agent \textit{P(wa)}, perspective of the agent on the theme \textit{P(at)}, effect on the theme \textit{E(t)} and agent \textit{E(a)}, value of the theme \textit{V(t)} and agent \textit{V(a)}, mental state of the theme \textit{S(t)} and agent \textit{S(a)}, power, and agency.

\subsection{Hyperparameters}
All models are trained with hidden size $H = 150$, number of definition words $N = 42$, number of related words $|R_{w^t}| = 20$ and dropout of $0.5$ to prevent overfitting. For emotion prediction we set $\theta = 0.5$. We use Concept-Net numberbatch embeddings ~\cite{numberbatch} because we find empirically that these outperform other pre-trained embeddings (GloVe and dependency-based embeddings~\cite{levy-goldberg-2014-dependency}) on the development set.

We tune our only hyperparameters on the development set: the weights $\lambda_a$ for the contribution of each loss term $\mathcal{L}^a$ to the total loss $\sum_{a} \lambda_a \mathcal{L}^a$ (see \ref{sec:learning}). We experiment with $10$ manually selected weight combinations, where each $\lambda_a \in (0, 5)$. We find that the optimal weights are: 
\begin{itemize}
    \item $\lambda_a = 0.3$ for $a \in \{SocialVal, Impact$, $V(t), V(a), power, agency\}$
    \item $\lambda_a = 0.167$ for $a \in \{Fact, Sent\}$
    \item $\lambda_a = 1.0$ for $a \in \{P(wt), P(wa), P(at)$, $E(t)$, $E(a)$, $S(t), S(a)\}$
    \item $\lambda_a = 0.5$ for $a = Polite$
    \item $\lambda_a = 3.0$ for $a = Emo$
\end{itemize}
Additionally, we compute expected validation performance~\citep{Dodge2019ShowYW} on each connotation aspect individually (see Table \ref{tab:appexpvalconn}).
% \begin{wraptable}{r}{0.5\textwidth}
\begin{table}[t]
\centering
\begin{tabular}{|l|ll|}
\hline
 & Best Dev & $\mathbb{E}[$Dev$]$ \\ \hline
 \hline
\begin{tabular}[c]{@{}l@{}}\textit{Social}\\ \textit{Val}\end{tabular}& 0.681 & 0.700 \\ %\hline
\textit{Polite} & 0.540 & 0.554 \\ %\hline
\textit{Impact}  & 0.704 & 0.711\\ %\hline
\textit{Fact} & 0.546 & 0.565\\ %\hline
\textit{Sent}& 0.612 & 0.631\\ %\hline
\textit{Emo}& 0.574 & 0.584 \\ \hline
\textbf{\textit{Avg}} & 0.610 & 0.610 \\ \hline
\hline
\textit{P(wt)} & 0.525 & 0.583\\ %\hline
\textit{P(wa)} & 0.580 &  0.577\\ %\hline
\textit{P(at)} & 0.606 &  0.613\\ %\hline
\textit{E(t)} & 0.668 &  0.686\\ %\hline
\textit{E(a)} & 0.596 & 0.618\\ %\hline
\textit{V(t)} & 0.376  & 0.469\\ %\hline
\textit{V(a)} & 0.463 & 0.459\\ %\hline
\textit{S(t)} & 0.600 & 0.642\\ %\hline
\textit{S(a)} & 0.611 &  0.603\\ %\hline
\textit{power} & 0.466 & 0.493\\ %\hline
\textit{agency} & 0.426 & 0.487\\ \hline
\textbf{\textit{Avg}} & 0.538 & 0.542\\ \hline
\hline
\textbf{\textit{Avg}} & 0.563 & 0.563\\ \hline
\end{tabular}
\caption{Best macro-averaged F1 on the development set and expected validation score~\cite{Dodge2019ShowYW} for our connotation representation learning model on all connotation aspects.}\label{tab:appexpvalconn}
\end{table}

\subsection{Training}
We optimize using Adam~\cite{KingmaAdam} with learning rate $0.001$ and minibatch-size of $64$ for $80$ epochs with early stopping. We optimize the parameters $W^a$, $b^a$ for each noun and adjective aspect $a$ separately from the parameters for each verb aspect $a$, allowing both to update the parameters of the definition encoder, and attention layer. 

\subsection{Detailed results}
We present aspect-level results for the task of connotation label prediction (see Table\ref{tab:detailresults}).
% \begin{wraptable}{r}{0.5\textwidth}
\begin{table}[t]
\centering
\begin{tabular}{|l|lllll|}
\hline
 & \textbf{Maj} & \textbf{LR} & \begin{tabular}[c]{@{}l@{}}\textbf{CE+R}\\ \textbf{(S)}\end{tabular} & \begin{tabular}[c]{@{}l@{}}\textbf{CE}\\ \textbf{(J)}\end{tabular}  & \begin{tabular}[c]{@{}l@{}}\textbf{CE+R}\\ \textbf{(J)}\end{tabular} \\ \hline
 \hline
\begin{tabular}[c]{@{}l@{}}\textit{Social}\\ \textit{Val}\end{tabular}& .228 & \uline{.664} & .651 &.632 & \textbf{.651}\\ %\hline
\textit{Polite} & .311 & .470 & .467 & \uline{\textbf{.518}} &  .464\\ %\hline
\textit{Impact}  & .278 & .669 & .681 & .687  & \uline{\textbf{.704}}\\ %\hline
\textit{Fact} & .271 & \uline{.576} & .531 & .549 &  \textbf{.560}\\ %\hline
\textit{Sent}& .247 & .585 & .606 & \uline{\textbf{.615}} &  \uline{\textbf{.615}} \\ %\hline
\textit{Emo}& .487 & \uline{.604} & .599 & .578 & \textbf{.587}\\ \hline
\textbf{\textit{Avg}} & .304 & .594 & .589 & .597 & .597  \\ \hline
\hline
\textit{P(wt)} & .246 & \uline{.501} & .437 & \textbf{.481} &  .439 \\ %\hline
\textit{P(wa)} & .213 &  .564 & .487 & .544 &  \uline{\textbf{.583}}\\ %\hline
\textit{P(at)} & .204 &  \uline{.649} & .553 &  .623 & \textbf{.629}\\ %\hline
\textit{E(t)} & .156 &  \uline{.721} & .673 & .655 & \textbf{.661}\\ %\hline
\textit{E(a)} & .226 & \uline{.573}  & .420 & \textbf{.557} & .530\\ %\hline
\textit{V(t)} & .109 &  .369 & .365 & \uline{\textbf{.391}} &  .373\\ %\hline
\textit{V(a)} & .320 &  \uline{.449} & .428 & \textbf{.375} &  .370 \\ %\hline
\textit{S(t)} & .286 & \uline{.640}  & .548 & .586 & \textbf{.629}\\ %\hline
\textit{S(a)} & .203 &  \uline{.551} & .481 & \textbf{.543} &  .537\\ %\hline

\textit{power} & .294 & .476 & .467 & .474  & \uline{\textbf{.480}}\\ %\hline
\textit{agency} & .182 & \uline{.589} & .515 &  \textbf{.505} & .490\\ \hline
\textbf{\textit{Avg}} & .222 & .553 & .489 & .521  & .520 \\ \hline
\hline
\textbf{\textit{Avg}} & .251 & .568 & .524 & .548 & .547 \\ \hline
\end{tabular}
\caption{Macro-averaged F1 results for connotation prediction on the test set. The upper part shows noun/adjective aspect results, the bottom shows verb aspect results. 
\uline{Underline} indicates the best performing model per row. \textbf{Bold} indicates the best performing joint learning model per row.}\label{tab:detailresults}
\end{table}
% \vspace{-10pt}
% \end{wraptable}

\section{Extrinsic Stance Evaluation}
\label{app:stance}
\begin{table}[t]
\centering
\begin{tabular}{|l|l|l|l|l|}
\hline
\textbf{Topic} & \textbf{\# Ex} & \textbf{\# C} & \textbf{\# P} & \textbf{\# N} \\ \hline
\hline
abortion & 12453 & 3962 & 5236 & 3255 \\ \hline
gay marriage & 11037 & 2907 & 5082 & 3048 \\ \hline
gun control & 10119 & 4610 & 2681 & 2828 \\ \hline
evolution & 9896 & 2586 & 4480 & 2830 \\ \hline
\begin{tabular}[c]{@{}l@{}}existence of\\ God\end{tabular} & 7227 & 2588 & 2517 & 2122 \\ \hline
\hline
death penalty & 2834 & 995 & 951 & 888 \\ \hline
\begin{tabular}[c]{@{}l@{}}humans are \\ responsible\end{tabular} & 1608 & 560 & 538 & 510 \\ \hline
\begin{tabular}[c]{@{}l@{}}marijuana \\ legalization\end{tabular}
& 1491 & 328 & 697 & 466 \\ \hline
\begin{tabular}[c]{@{}l@{}}communism \\is better than\\ capitalism\end{tabular} & 1279 & 618 & 277 & 384 \\ \hline
\hline
\begin{tabular}[c]{@{}l@{}} illegal\\ immigration\end{tabular}& 291 & 108 & 87 & 96 \\ \hline
\begin{tabular}[c]{@{}l@{}}health care\\ reform \end{tabular}& 201 & 76 & 51 & 74 \\ \hline
\begin{tabular}[c]{@{}l@{}}legalize\\ prostitution\end{tabular} & 199 & 57 & 88 & 54 \\ \hline
Israel & 100 & 29 & 38 & 33 \\ \hline
\begin{tabular}[c]{@{}l@{}}vegetarian \\diet is best\end{tabular} & 79 & 29 & 29 & 21 \\ \hline
\begin{tabular}[c]{@{}l@{}}women in \\combat\end{tabular} & 47 & 15 & 19 & 13 \\ \hline
\begin{tabular}[c]{@{}l@{}}minimum\\ wage \end{tabular}& 27 & 9 & 8 & 10 \\ \hline
\hline
\textbf{Overall} & 58888 & 19477 & 22779 & 16632\\ \hline
\end{tabular}
\caption{Statistics for the stance detection dataset. C indicates `con', P indicates `pro', N indicates 'neutral'.}\label{tab:stancestat}
\end{table}
\subsection{Dataset Details} We map the topic-stance annotations in the Internet Argument Corpus to individual topics and labels (e.g., `pro-life' $\rightarrow$ topic `abortion' with label `con'). We show dataset statistics in Table \ref{tab:stancestat}, where topics in the upper part are large sized, topics in the middle part are medium sized, and topics in the lower part are small sized.

\subsection{Training Details} We split the data $60\%$ train, $20\%$ development, and $20\%$ test. We train our models using pre-trained 100-dimensional word embeddings from GloVe~\cite{glove}, as these are comparable to and more time-efficient than larger word embeddings. We use a hidden size of $60$, dropout of $0.5$, and train for $70$ epochs with early stopping on the development set. We optimize Adam with learning rate $0.001$ and minibatch-size of $64$ on the cross-entropy loss. Our hyperparameters are set following~\citet{Augenstein2016StanceDW}.  

When truncating to medium size in \S \ref{sec:exteval}, we truncate train topics to at most $2000$ examples  (\texttt{Trunc Train} and \texttt{Trunc All}) and development and test topics to at most $600$ examples (\texttt{Trunc All}). 

\subsection{Topic Stance Analysis} We present a detailed analysis of the results of the models \texttt{BiC+W} and \texttt{BiC+C} on the stance detection on individual topics. First, we find that when the models are trained with all of the data (All Data), there are statistically significant differences on only two topics, one of which is very small (see Table \ref{tab:appstance1}). This is further evidence that the models are comparable in this setting.

We then find that when trained with truncated training data (see \S \ref{sec:exteval} for details) (\texttt{Trunc Train}), \texttt{BiC+C} improves over \texttt{BiC+W} on six topics, including four of the medium or truncated large topics (see Table \ref{tab:appstance2}). When trained and evaluated with truncated data (\texttt{Trunc All}), \texttt{BiC+W} and \texttt{BiC+C} have statistically significant improvements over each other on the same number of topics (two each) but \texttt{BiC+C} is significantly better overall (see Table \ref{tab:appstance3}). These results further show that connotations help to learn stance when data is limited.

\begin{table*}
\begin{subtable}{0.33\textwidth}
    \centering
    \begin{tabular}{|l|l|l|}
        \hline
        \textbf{Topic} & \begin{tabular}[c]{@{}l@{}}\textbf{\texttt{BiC}}\\ \textbf{\texttt{ +W}}\end{tabular} & \begin{tabular}[c]{@{}l@{}}\textbf{\texttt{BiC}}\\ \textbf{\texttt{ +C}}\end{tabular} \\ \hline
        \hline
        \textit{abortion} & .49 & .49 \\ \hline
        \begin{tabular}[c]{@{}l@{}}\textit{gay}\\\textit{marriage}\end{tabular} & .47 & .48 \\ \hline
        \textit{gun control} & .53 & .55 \\ \hline
        \textit{evolution }& .43 & .44 \\ \hline
        \begin{tabular}[c]{@{}l@{}}\textit{existence} \\ \textit{of God}\end{tabular} & .52 & .52 \\ \hline
        \hline
        \begin{tabular}[c]{@{}l@{}}\textit{death} \\\textit{penalty}\end{tabular} & .45 & .50$^{\dagger}$ \\ \hline
        \begin{tabular}[c]{@{}l@{}}\textit{humans are }\\\textit{ responsible}\end{tabular} & .49 & .54 \\ \hline
        \begin{tabular}[c]{@{}l@{}}\textit{marijuana}\\ \textit{legalization}\end{tabular} & .51 & .50 \\ \hline
        \begin{tabular}[c]{@{}l@{}}\textit{communism}\\ \textit{ is better than }\\ \textit{capitalism}\end{tabular} & .52 & .54 \\ \hline
        \hline
        \begin{tabular}[c]{@{}l@{}}\textit{illegal }\\ \textit{immigration}\end{tabular} & .34 & .38 \\ \hline
        \begin{tabular}[c]{@{}l@{}}\textit{health care}\\\textit{ reform}\end{tabular} & .64 & .91 \\ \hline
        \begin{tabular}[c]{@{}l@{}}\textit{legalize }\\\textit{ prostitution}\end{tabular} & .40 & .53 \\ \hline
        \textit{Israel }& .66 & .42 \\ \hline
        \begin{tabular}[c]{@{}l@{}}\textit{vegetarian} \\\textit{ diet is best}\end{tabular} & .52$^{\dagger}$ & .33 \\ \hline
        \begin{tabular}[c]{@{}l@{}}\textit{women in }\\ \textit{combat}\end{tabular} & .28 & .30 \\ \hline
        \begin{tabular}[c]{@{}l@{}}\textit{minimum} \\ \textit{wage}\end{tabular} & .30 & .22 \\ \hline
        \hline
        \textbf{Overall} & .57 & .57\\ \hline
    \end{tabular}
    % \caption{Macro F1 results on the test set when training in scenario `All Data'.$^\dagger$ indicates significane with $p < 0.01$.}\label{tab:appstance1}
    \caption{On \texttt{All Data}.}\label{tab:appstance1}
    \vspace{8pt}
\end{subtable}%
\begin{subtable}{.33\textwidth}
    \centering
    \begin{tabular}{|l|l|l|}
        \hline
        \textbf{Topic} & \begin{tabular}[c]{@{}l@{}}\textbf{\texttt{BiC}}\\ \textbf{\texttt{ +W}}\end{tabular} & \begin{tabular}[c]{@{}l@{}}\textbf{\texttt{BiC}}\\ \textbf{\texttt{ +C}}\end{tabular} \\ \hline
        \hline
        \textit{abortion} & .46 & .47 \\ \hline
        \begin{tabular}[c]{@{}l@{}}\textit{gay}\\\textit{marriage}\end{tabular} & .48 & .46 \\ \hline
        \textit{gun control} & .50 & .55$^*$ \\ \hline
        \textit{evolution} & .41 & .43$^{*\dagger}$ \\ \hline
        \begin{tabular}[c]{@{}l@{}}\textit{existence} \\ \textit{of God}\end{tabular} & .48 & .51$^*$ \\ \hline
        \hline
        \begin{tabular}[c]{@{}l@{}}\textit{death} \\\textit{penalty}\end{tabular} & .48 & .50 \\ \hline
        \begin{tabular}[c]{@{}l@{}}\textit{humans are }\\\textit{ responsible}\end{tabular} & .45 & .53$^{*\dagger}$ \\ \hline
        \begin{tabular}[c]{@{}l@{}}\textit{marijuana}\\ \textit{legalization}\end{tabular} & .51 & .50 \\ \hline
        \begin{tabular}[c]{@{}l@{}}\textit{communism}\\  \textit{is better than} \\ \textit{capitalism}\end{tabular} & .53 & .54 \\ \hline
        \hline
        \begin{tabular}[c]{@{}l@{}}\textit{illegal }\\ \textit{immigration}\end{tabular} & .45$^*$ & .36 \\ \hline
        \begin{tabular}[c]{@{}l@{}}\textit{health care}\\ \textit{reform}\end{tabular} & .62 & .64$^{*\dagger}$ \\ \hline
        \begin{tabular}[c]{@{}l@{}}\textit{legalize }\\\textit{ prostitution}\end{tabular} & .49 & .50 \\ \hline
        \textit{Israel} & .54 & .44 \\ \hline
        \begin{tabular}[c]{@{}l@{}}\textit{vegetarian} \\\textit{ diet is best}\end{tabular} & .10 & .11$^*$ \\ \hline
        \begin{tabular}[c]{@{}l@{}}\textit{women in }\\\textit{ combat}\end{tabular} & .52 & .36 \\ \hline
        \begin{tabular}[c]{@{}l@{}}\textit{minimum }\\ \textit{wage}\end{tabular} & .22 & .33 \\ \hline
        \hline
        \textbf{Overall} & .54 & .56$^{*\dagger}$\\ \hline
    \end{tabular}
    \caption{On \texttt{Trunc Train}.}\label{tab:appstance2}
    \vspace{8pt}
    % \caption{Macro F1 results on the test set when training in scenario `Trunc Train'. $^*$ indicates significance with $p < 0.05$, $^\dagger$ indicates significane with $p < 0.01$.}
\end{subtable}%
\begin{subtable}{0.33\textwidth}
% % \begin{table}[t]
    \centering
    \begin{tabular}{|l|l|l|}
        \hline
        \textbf{Topic} & \begin{tabular}[c]{@{}l@{}}\textbf{\texttt{BiC}}\\ \textbf{\texttt{ +W}}\end{tabular} & \begin{tabular}[c]{@{}l@{}}\textbf{\texttt{BiC}}\\ \textbf{\texttt{ +C}}\end{tabular} \\ \hline
        \hline
        \textit{abortion} & .49 & .48 \\ \hline
        \begin{tabular}[c]{@{}l@{}}\textit{gay}\\\textit{marriage}\end{tabular} & .49$^*$ & .46 \\ \hline
        \textit{gun control} & .51 & .50 \\ \hline
        \textit{evolution} & .43 & .43 \\ \hline
        \begin{tabular}[c]{@{}l@{}}\textit{existence} \\\textit{ of God}\end{tabular} & .49 & .47 \\ \hline
        \hline
        \begin{tabular}[c]{@{}l@{}}\textit{death} \\\textit{penalty}\end{tabular} & .48 & .46 \\ \hline
        \begin{tabular}[c]{@{}l@{}}\textit{humans are }\\\textit{ responsible}\end{tabular} & .46 & .55$^{*\dagger}$ \\ \hline
        \begin{tabular}[c]{@{}l@{}}\textit{marijuana}\\ \textit{legalization}\end{tabular} & .50 & .49 \\ \hline
        \begin{tabular}[c]{@{}l@{}}\textit{communism}\\ \textit{ is better than }\\ \textit{capitalism}\end{tabular} & .55 & .52 \\ \hline
        \hline
        \begin{tabular}[c]{@{}l@{}}\textit{illegal }\\ \textit{immigration}\end{tabular} & .44 & .44 \\ \hline
        \begin{tabular}[c]{@{}l@{}}\textit{health care}\\textit{\ reform}\end{tabular} & .54 & .64$^{*\dagger}$ \\ \hline
        \begin{tabular}[c]{@{}l@{}}\textit{legalize }\\\textit{ prostitution}\end{tabular} & .47 & .51 \\ \hline
        Israel & .41 & .56 \\ \hline
        \begin{tabular}[c]{@{}l@{}}\textit{vegetarian} \\\textit{ diet is best}\end{tabular} & .52$^{*\dagger}$ & .50 \\ \hline
        \begin{tabular}[c]{@{}l@{}}\textit{women in }\\\textit{ combat}\end{tabular} & .43 & .47 \\ \hline
        \begin{tabular}[c]{@{}l@{}}\textit{minimum }\\ \textit{wage}\end{tabular} & .33 & .35 \\ \hline
        \hline
        \textbf{Overall} & .54 & .56$^{*\dagger}$\\ \hline
    \end{tabular}
    % \caption{Macro F1 results on the test set when training in scenario `Trunc All'. $^*$ indicates significance with $p < 0.05$, $^\dagger$ indicates significane with $p < 0.01$.}
    \caption{On \texttt{Trunc All}.}\label{tab:appstance3}
    \vspace{8pt}
\end{subtable}
\caption{Macro F1 results on the test set for three different data scenarios. $^*$ indicates significance with $p < 0.05$, $^\dagger$ indicates significane with $p < 0.01$.}
\end{table*}

\end{document}